# Approaches to Validation of Information Fusion Systems

Alexander Kott
Army Research Laboratory
Adelphi, MD, U.S.A
alexander.kott1@us.army.mil

Wes Milks
Lockheed Martin Corporation
Orlando, FL, U.S.A.
wesley.milks@lmco.com

*Abstract* – We motivate and offer a formal definition of validation as it applies to information fusion systems. Common definitions of validation compare the actual state of the world with that derived by the fusion process. This definition conflates properties of the fusion system with properties of systems that intervene between the world and the fusion system. We propose an alternative definition where validation of an information fusion system references a standard fusion device, such as recognized human experts. We illustrate the approach by describing the validation process implemented in RAID, a program conducted by DARPA and focused on information fusion in adversarial, deceptive environments.

## 1 Introduction

The concept and term *validation* appear infrequently in the fusion literature [1, 2]. A rigorous definition of validation as it applies to information fusion does not seem to exist. On the other hand, the field of modeling and simulation does offer extensive literature focused on validation, and this may be a good starting point for our discussion. In [3], validation is defined as a process that determines the degree to which a model or simulation is an accurate representation of the real world from the perspective of the intended uses of the model or simulation.

Consider how one can apply this definition to information fusion. As illustrated in Fig. 1, the fusion function $F$ (a system that performs fusion) receives streams of information $\{r\}$ from observer-reporter $R$ (often consisting of sensors, human observers, etc.) and transforms these inputs into an estimate of the true state of the observed world $W$. In many practical cases, the world includes elements that wish to conceal or misrepresent their actions and intent; these elements—let us call them the deceiver $D$—attempt to disguise or corrupt some of the information available to the observer-reporter $R$.

Validation of a given fusion system $F$, then, can be defined as a (usually experimental) proof that $F$ is likely (in a probabilistic sense) to estimate the true state $S$ of the world $W$ with an error that does not exceed an acceptable value $\delta$:

$$P(|S\text{-}S'| < \delta) > \theta, \qquad (1)$$

where $S'=F(\{r\})=F(R(D(S)))$ is the estimate produced by the fusion system $F$; $S$ is the true state of the world $W$, $\delta$ is the maximum acceptable error, and $\theta$ is the minimum acceptable probability that the estimate does not exceed $\delta$. Here $|S\text{-}S'|$ is a domain-specific metric that quantifies the differences between the two states. Later in this paper, we bring practical examples of such metrics. Henceforth, we call the approach defined by Fig. 1 and Equation (1) Validation Type 1.

However, we find this definition unsatisfactory in practice. Because the error $|S\text{-}S'|$ depends on the competency of observer $R$ and deceiver $D$ as much as on efficacy of fusion $F$, a fusion system that is well validated for the given $R$ and $D$ may be invalid for a different $R$ or different $D$. In many practical environments, $R$ and especially $D$ vary widely—we will give an example of such an environment shortly. In other words, the above definition of validation conflates the efficacy of the fusion process $F$ with craftiness of $D$ or inadequacy of $R$. This is not a helpful characterization of $F$.

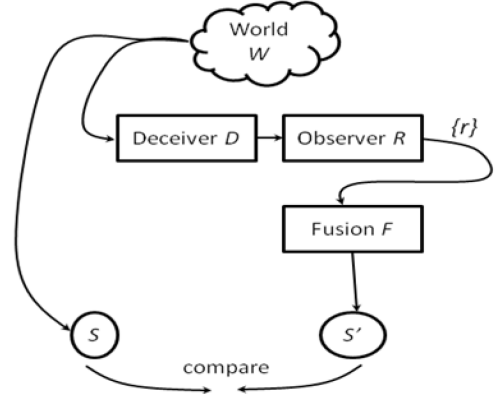

Figure 1. Validation by comparing the real-world state with the state estimated by the fusion system.

Therefore, we offer a different definition. Assume we have a fusion device $G$ accepted as a standard of performance in fusion. Then, we can define the validation of a given fusion system $F$ as a proof (Fig. 2) that $F$ is likely to estimate the true state $S$ of the world W with an error no greater than the error produced by the standard fusion device $G$:

$$P(|S\text{-}S'|<|S\text{-}S''|) > \vartheta, \qquad (2)$$

where $S'=F(\{r\})$ is the estimate produced by fusion system $F$, $S''$ is the estimate produced by the standard fusion device $G$, $S$ is the true state of the world $W$, and $\vartheta$ is the minimum acceptable probability that the error of $S'$ is within the bounds of the standard device's error.

Henceforth, we call the approach defined by Fig. 2 and Equation (2) Validation Type 2.

The key advantage of the Type 2 approach is that it mitigates the conflation of fusion properties with those of $R$ and $D$. Indeed, consider two deceivers $D1$ and $D2$ of greatly different effectiveness. In Validation Type 1, the apparent fusion error $|S-S'_{D1}|$ may differ greatly from $|S-S'_{D2}|$. Not so in Validation Type 2. Under a reasonable assumption that for all $S$ and $D$ of interest $|F(D(S))-G(D(S))| < \delta$, it can be readily shown that $|S-S'|-|S-S''|$ remains relatively constant (within $\delta$) regardless of $D$.

But where does one find a "standard fusion device" with which to compare a given $F$ for the purposes of Validation Type 2? In practice, we argue, one can use human experts or a legacy fusion system as a standard of comparison. In the remainder of the paper, we describe a particular case in which we implemented this validation approach.

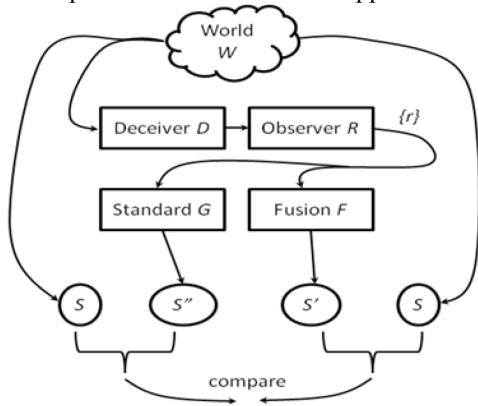

Figure 2. Validation by comparing the estimation error of the fusion system with that of an accepted standard.

## 2 An Illustrative Problem

The validation work described in this paper was a part of the research program titled Real-time Adversarial Intelligence and Decision-making (RAID) sponsored by the United States' Defense Advanced Research Programs Agency (DARPA) during the period of 2004-2008 [4]. The objective of the program was to build tools for fusion in military operations. The tools were to estimate in real-time the enemy's likely goals, deceptions, actions, movements and positions. Although many types of military operations can benefit from such capabilities, the RAID program focused on an intentionally narrow but still very challenging domain: the in-execution, tactical combat of infantry (supported by armor and air platforms) against a guerilla-like enemy force in an urban terrain. (In the following discussion, we adhere to the US military convention: the military force that uses the RAID system is called the friendly or "Blue" force, while their opponent is called the enemy or "Red" force.)

In a typical problem addressed by RAID, a largely infantry Blue Force operates in an urban terrain against an insurgent-like irregular infantry Red Force. Real-world examples of such military engagements include Mogadishu [5] and Fallujah [6]. The problem may involve the defense of Blue facilities, the rescue of downed aircrew, the capture of an insurgent leader, the rescue of hostages or the reaction to an attack on a Blue patrol.

To illustrate such a problem, consider a scenario where Blue forces receive the task to protect four buildings. The Red's objective was to attack any number of those four facilities to inflict casualties and damage. The scenario left both forces with a very broad range of tactical options. For example, Blue forces could attempt to move quickly into defensive positions around the designated buildings. Some Red forces were initially closer to the buildings of interest and could cause damage and casualties while Blue forces were enroute. The success of the Blue force's mission was influenced in particular by how quickly Blue forces were in place around the attacked buildings and how much (or how little) fire the buildings received from the Red forces. The Red forces could attack all the target buildings, or focus their efforts on only one or two of them. The Blue commander could try to protect all the buildings, or leave unprotected a single building to protect better the remaining buildings. In most cases, the Red force would attempt to deceive the Blue force by concealment and other means [7].

In an illustrative problem like this, the Blue force could be a company comprised of 10-30 infantry fire-teams (each typically with four foot soldiers armed with rifles and light machine guns) and 2-8 armored attack vehicles, supported by artillery or attack helicopters. The Red forces could consist of 20-40 teams (each typically consisting of about three foot fighters armed with assault rifles, rocket propelled grenade launchers, machine guns and remotely activated bombs. The battle area could be about 10-30 square kilometers with multiple single-story or multistory buildings. The time scale of such a battle could be about two to twelve hours.

## 3 Fusion Input and Output

In planning and executing a battle like this, the company commander, his supporting staff (including possibly the staff at the higher echelon of command) and his subordinate unit leaders would receive and fuse (mentally or with the aid of computerized fusion system like RAID) a bewildering array of information [8]. These are the $\{r\}$ in Fig 1 and 2. Some of this information is relatively immune to adversarial manipulation. For example, information on the Blue force composition and mission plan; detailed maps of the area, potentially including detailed 3D data of the urban area; known concentrations of non-combatants such as markets; culturally sensitive areas such as worship houses; reports of historic and recent prior activities such as explosions of roadside bombs in the area; continuous updates on the locations and status of the Blue force as they move before and during the battle.

Other input information would be highly incomplete, often erroneous, and potentially distorted by the Red concealment and deception. For example, Blue forces' reports (electronic, verbal or textual) regarding the observed positions and strength of the Red force, or fires received from the Red force; prior information about the Red force's tactics, and preferences; estimates of the Red force locations, strengths, equipment, probable mission, morale, competency, intent and plan; assessments of outcomes of alternative battle plans for Red and Blue forces. Much of this input information is highly dynamic: it arrives and changes as the battle unfolds.

As far as the output of the fusion process, i.e., the *S'* in Fig. 1 and 2, users typically desire two types of information. First is the estimate of the Red force's current situation: estimated actual locations of the Red force (most of which are normally concealed and are not observed by the Blue force); the current intent of the Red force, and potential deceptions that the Red force may be performing. Generation of such an estimate is often called Level-2 fusion [9].

The second type of fused information describes the estimated future events: Red force's future locations (as a function of time), movements, fire engagements with the Blue force, changes in strengths and intent. Generation of such an estimate is often called Level-3 fusion [9]. It may also include recommendations to the Blue force regarding the suitable ways to prevent or to parry such Red force actions.

The RAID system as a whole performs both Level-2 and Level-3 fusion. However, this paper focuses on only one aspect of the RAID effort that falls into the Level-2 fusion area. It is important to stress the adversarial nature of this fusion process: the correctness of the fused output can suffer if the Red force intentionally distorts the input information. By cleverly concealing some actual information while creating and exposing false information, the Red force can lead the Blue force's fusion process toward erroneous (and at least partially predictable to the Red force) conclusions. The use of such a misleading fused product can be catastrophic for the Blue force.

## 4 Overview of the Fusion System

The RAID system consists of two major components, the Adversarial Reasoning Module (ARM) and the Deception Reasoning Module (DRM). In addition, the Combat Simulation System simulates the events occurring on the battlefield.

There are two major inputs to the RAID system, battlefield models and situational information. The battlefield models include internal models of the terrain (3-D terrain data to include doors, windows, floors, and other details of the landscape); internal models of Red and Blue resources (weapons parameters to include range, lethality, probabilities of acquisition and destruction, platform parameters to include minimum, maximum and average speeds, vulnerability to specific weapons); and internal models of actions (to include influence of goals and objectives (speed, safety, success) on the selection of tactics).

Situational information includes Red and Blue force strength, location, and movement and is comprised of intelligence data and updates from deployed sensors and operational data from both mission planning and operational reports from the deployed forces. The output of the RAID system includes estimates of Red current situation and future actions, and recommendations for the Blue force.

The DRM identifies the current Red situation (corresponding mainly to Level-2 fusion), including probable Red concealments and deceptions. The DRM provides the ARM with probable locations and concentrations of un-observed Red assets for use in the ARM calculations.

The ARM generates estimates of the future Red actions and recommendations for the Blue actions, either on-demand or in response to battle situation changes. As information regarding the battlefield situation (locations, strengths, postures, actions, etc.) of Red and Blue troops becomes available or changes, the ARM generates a new or modified set of estimates. In this paper, we focus on a subset of DRM and do not address ARM.

The Combat Simulation System was not a part of RAID proper, but served only the purposes of experimentation. It used the US Army simulation system called OneSAF Testbed Baseline (OTB) [10] with certain modifications to meet the needs of RAID experimentation. Both Blue and Red forces were controlled by specially trained human operators.

Fusion algorithms of RAID are beyond the scope of this paper and are discussed elsewhere [4, 11, 12, 13].

## 5 Validation Experiments

As discussed in the Introduction, if we were to judge the fusion accuracy of RAID by comparing its fused output with real-world state, the results would depend greatly on the effectiveness of the observers-reporters (Blue force) and on effectiveness of deceivers (Red force). The effectiveness of the observers-reporters and of deceivers, however, varies greatly depending on a battle situation. Thus, we elected to use the Validation Type 2 approach summarized in Equation (2).

Another consideration in our validation methodology was to include experimental cases of sufficient military realism and complexity, to validate the technology in reference to its intended military use. Ideally, real-world examples of military operation would provide specific data for experimental assessments. Unfortunately, we found very few suitable real-world cases where actual enemy situation data were available with sufficient fidelity and certainty. Live-force wargames, often very realistic [14], could be an alternative to real-world cases. However, introduction of new technologies in such live-force wargames is difficult and expensive, and it is rarely possible to control the content of the wargames to any significant extent.

These considerations led us to the following concept of experiment. Each experiment consisted of multiple wargames executed by live Red and Blue commanders but in a simulated computer wargaming environment. We selected a wargaming environment that the US military widely used and recognized as relatively realistic. In half of the wargames the Blue commander received the support of a human team of competent assistants (staff) who in effect played the role of the standard fusion device $G$ in Fig. 2. Their responsibilities included producing running estimates $S''$. This set of wargames constituted the control group. In the other half of wargames Blue commander operated without a human staff. Instead, he obtained a similar support from the RAID fusion system which produced $S'$. These wargames constituted the test group. The data collection and redaction process compared the accuracy of the control group with the accuracy of the test group, as in Fig. 2.

### 5.1 Validation Experiments' Structure

The Red cell (Fig.3) was a group of military experts, typically 5-6 retired officers specially trained to impersonate Red force in military wargames. The Red cell acted as the decision-makers of the Red force and commanded the simulated Red fighters. The Red cell operated in so called free-play mode (i.e., they were not constrained by a pre-conceived plan of actions) and emulated an intelligent, innovative, hard-to-anticipate adversary.

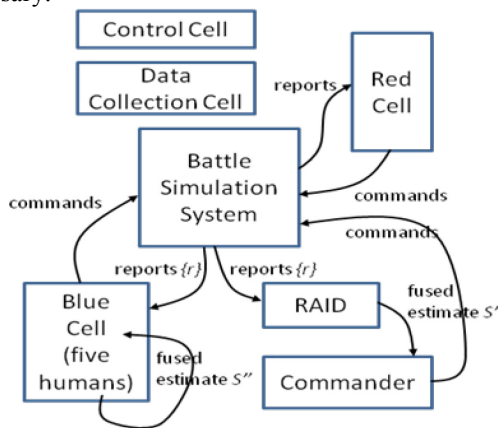

Figure 3. The structure of the validation experiments. Compare to Fig. 2.

The Red force's creative use of urban terrain, rapid movement in the familiar city, concealment, deceptions, ambushes, remotely controlled bombs, rocket-propelled grenades, heavy machine guns, infiltration and civilian spies challenged the simulated Blue forces and their human commanders. We physically separated the Red cell from the Blue cell, and took a number of precautions to prevent information exchange between the Red and Blue subjects. In particular, the Red cell did not know if a given wargame was a control wargame (with the human staff) or a test wargame (with the RAID system). Exit interviews with the Red cell confirmed that they were not able to discern whether their opponent was human or computer-assisted.

The Blue cell of control wargames was a commander supported by four assistants (staff). The Blue cell acted as the decision-makers of the Blue force and commanded the simulated Blue soldiers. Typically, the commander was a retired or active-duty officer of the US Army. Ranks of the officers varied from captain to lieutenant colonel. The ranks of the staff members varied from senior non-commissioned officers to majors.

The Blue cell of test wargames was a human commander supported by the RAID system, which acted as a surrogate staff. RAID received the same information that was available to the staff of the control wargames.

The simulation system executed the commands of the Blue and Red cells and simulated the movements, observations, fires and physical effects experienced by the opposing forces and by the civilians caught in the fire.

The control cell included 3-5 persons responsible for initiating the wargame and resolving occasional breakdowns and anomalies in the wargaming process. The data collection cell included 3-6 people who operated the automated data collection software.

### 5.2 Experimental Procedure

We conducted experiments approximately every 6 months during the years 2004-2006. Each experiment included 20-30 wargames executed over the period of about two weeks. Adding the time required for hardware and software installation, testing, and subject training, each experiment took approximately four weeks. Experiments included different versions of the RAID software as its maturity gradually increased. Locations of experiments varied, and most took place at military bases. The typical physical setting was a comfortable conference room, and there was no attempt to emulate the extreme physical and psychological stress factors that attend the real-world battle command. The wargames were randomly divided in two approximately comparable groups (control group and test group) based on the assessments of SME. In each experiment, several subjects (members of the Blue cell) acted as commanders. We randomized the allocation of a wargame to a commander and to a time period within the experiment.

Each wargame took between 90 and 150 minutes; there were 2-3 wargames per day. Each wargame differed from others by initial situation of the forces, missions, objectives, and other significant details.

At the start of a wargame, the control cell met with Blue and Red cell (separately), explained the military situation and the objectives (different and suitable to the Red and Blue perspectives), and gave them such partial information about their respective enemy that could be realistically available to each side through their respective means of military intelligence. For example, the control cell would provide the Blue cell with accurate observations on locations and strengths of about 20% of the Red forces. Following a brief period of planning and preparation (15-30 minutes), the Blue and Red forces would begin the battle.

During the battle of a control wargame, the Blue commander and his supporting staff continually received information $\{r\}$ from the simulation system. The simulation system provided the Blue cell with accurate and complete information on the location of Blue forces, and such partial and potentially inaccurate observations of the Red forces that would be realistically available to the soldiers and sensors of the Blue force.

Examples of observations *{r}* included seeing several enemy fighters with rocket launchers on the roof of a building 400 meters away, or receiving machine gun fire from a window of a building.

The Blue commander and staff collaboratively and continually formulated and updated their estimates of the enemy situation ("Where are they hiding? How strong are they? What are they going to do next?"). For data collection purposes, we asked the staff to keep their estimates *S"* depicted on a computer screen, which they tended to update every few minutes.

Staff also provided recommendations to the commander regarding the possible actions he could take in the unfolding battle. The nature and specific procedures of collaboration between the commander and the staff members varied depending on the style and preferences of a particular commander, as it would be the case in the real world.

Having made his decision, the commander would issue a command-via a simulated radio link–to his forces, e.g., for a particular platoon or squad to change the direction of its advance. The tempo of operations was high, situation on the battlefield changed rapidly, and the commander had to issue commands nearly non-stop, although lulls of 2-5 minutes were also common. The commands issued by the commander went to a special group of 4-5 operators who translated the military-style commands into clicks at the simulation system interface. Operations of the Red cell were roughly similar to the ones described above.

During the test wargames, the Blue cell included only the commander. The role of staff was assumed by the RAID fusion system, which received the same information *{r}* that would be otherwise provided to the Blue staff. RAID generated estimates *S'* of the enemy situation and displayed this information when the commander asked for it (with a single button click), which typically occurred every 5-15 minutes depending on the commander's style and the battle situation. RAID generated and presented an updated estimate to the commander within the time acceptable to the users (between 30 and 300 seconds [4]).

During the wargame, the data collection cell used both computerized and manual means to collect continually the relevant information. This included discussions and decisions of the Blue staff, estimates *S"* produced by the Blue staff (in control wargames), estimates *S'* produced by RAID (in the test wargames), commands of the Blue commander, events occurring on the battlefield, Red and Blue casualties, and locations of the Blue and Red forces, i.e., the actual state *S*.

The RAID estimates automatically entered a computerized experiment log, and included estimated location of each entity at 1-minute increments. Collecting estimate data from the human staff was more challenging. The staff required time to analyze and produce the estimate, and to translate the estimate onto the data collection form. To reduce the time to collect the data, we provide the staff members with an automated data collection tool. As the run progressed, the staff generates new estimates approximately every 15 minutes.

### 5.3 Experimental Scenarios

Each scenario represented a battle in an urban terrain where a largely dismounted Blue Force operated against an insurgent-like irregular dismounted Red Force. The scenarios drew inspiration and realistic details from actual military engagements in Iraq in 2003-2006, and other urban battles. The situations included the defense of Blue government facilities, the rescue of downed aircrew, the capture of an insurgent leader, the rescue of hostages, or the reaction to an attack on a Blue patrol. For the sake of simplicity, we used only a few mission types. These were Point/Area Attack (Blue forces attack a specific target or area), Point/Area Defense (Blue forces protect specific buildings and clear all Red forces from a specific area) and Withdrawal (all Blue forces congregate at a designated location).

In a typical scenario, the Blue forces consisted of 18 teams (each with four dismounted soldiers armed with assault rifles and light machine guns) and five armored attack vehicles. For battle control purposes, the 18 teams operated as three platoons of six teams each with one simulation operator controlling each platoon. The Red forces consisted of 20 teams (each with three dismounted fighter armed with assault rifles and rocket propelled grenade launchers). Being more familiar with the terrain, Red forces moved somewhat faster than the Blue force. The battle area was a 2-kilometer (Km) by 2 Km region of a city with numerous multistory buildings. By experiment design, each scenario took about two hours to complete the mission. Prior to the wargame, both the Red and the Blue team received partial information regarding the opponent, such as location, strength, and possible intent of the opposing forces. The information typically covered only about 20-25% of the opposing force.

### 5.4 Metrics

In order to compute $|S\text{-}S'|$ and $|S\text{-}S''|$ in Equation (2), we had to develop a suitable metric that objectively measures the accuracy of estimated Red locations and strengths. One can view a running estimate of Red situation as a distribution of Red strength over the battlefield area. Experiments provide us with three such distributions (i.e., *S, S', S"* in Equation 2): one is the actual locations and strengths of the Red force recorded in the simulation log; the second is the estimate produced by the human staff, and the third is the estimate produced by RAID. If we have a metric that characterizes the distance between two distributions-specifically RAID-vs-actual and staff-vs-actual–we could determine whether staff's estimate or RAID's estimate is closer to the actual Red situation.

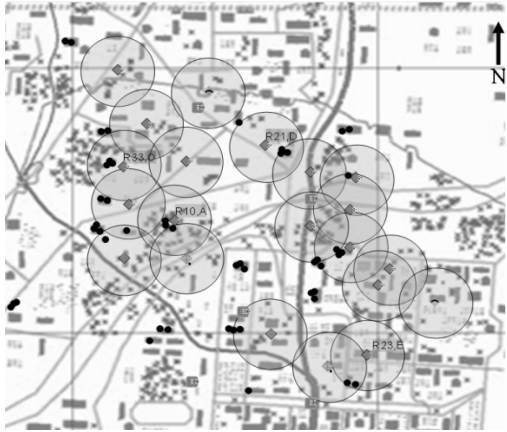

Figure 4. A CEP-like metric for comparing the estimates produced by the RAID fusion system with real world.

A common metric for measuring the distance between two distributions is the Prohorov Metric [15]. Unfortunately, we found the Prohorov metric computationally intractable for our problem. Other common metrics are the $L_p$ norms that provide a measure of the error (or distance) between two functions. As $p$ increases, the metric increases the emphasis on large errors at single locations over small errors distributed across the full metric space.

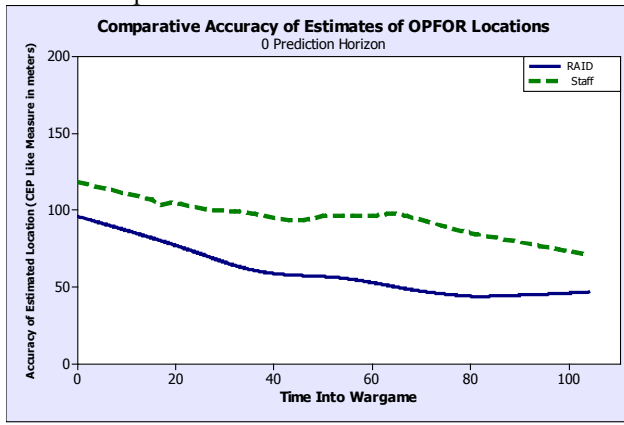

Figure 5. Plot of the estimate accuracy as a function of the time into the wargame. Smaller value of CEP measure indicates greater accuracy of estimates.

Unfortunately, we found the $Lp$ norms have a major disadvantage: SME do not perceive them as domain-relevant or intuitive. To provide domain relevance, we sought a physical measure of the estimate accuracy. We noted that military community uses Circular Error Probable (CEP) as a measure of the precision of a weapons system [16]. CEP is the radius of a circle into which a projectile will land 50% of the time, assuming the center of the circle is the actual aim point.

For the purposes of RAID experiments, we created a CEP-analogue for location estimate accuracy. Imagine drawing circles centered at all estimated locations and then enlarging the radius of each circle until 50% of the actual locations are contained within the circles (see Fig. 4). The result was a physical measure (in meters) that provides a domain relevant measure of the closeness between the estimated and actual locations. Although well received by domain experts, the CEP measure does have its drawbacks. It is not a true metric and also does not scale well, i.e., the measure only considers the best half of the estimated distribution, moving to measure that considered a 60% overlap could lead to significantly different conclusions.

### 5.5 Comparing Errors of Fused Estimates

With the CEP-like metric, we can use our experimental data to determine the values of $|S-S'|$, $|S-S''|$ and $P(|S-S'|<|S-S''|)$ of Equation (2).

Fig. 5 shows a comparison of the RAID-produced estimates $S'$ and staff-produced estimates $S''$ of Red current locations and strengths: CEP measure as a function of the time since the wargame started. For the CEP measure, smaller is better. The line is a Lowess smoother [17] through the respective RAID and staff data points. As would be intuitively expected, both the RAID and staff estimates get better as the wargame proceeds.

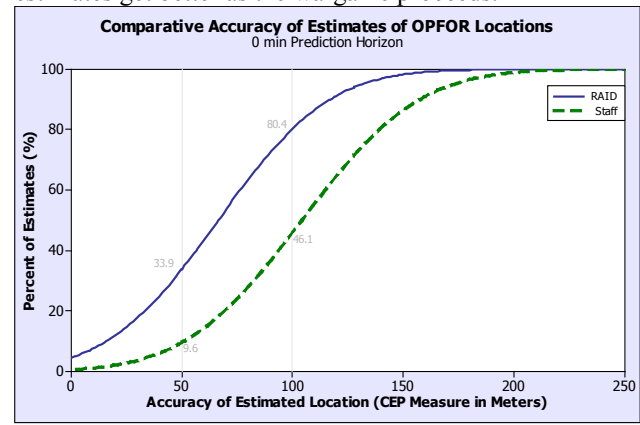

Figure 6. The cumulative probability on the y-axis for a range of estimate accuracy measures shown on the x-axis. In this plot, the higher line indicates better estimate.

Cumulative probability plots provide a graphical display of the estimated probability regression equation. The example in Fig. 6 displays the cumulative probability on the y-axis for a range of estimate accuracy measures shown on the x-axis. In this plot, the higher line indicates better estimate accuracy measures.

Reading the plot across from the y-axis, one can see the expected estimate accuracy measure for a given fraction percent of the estimates. For example, reading across from 60% until each regression line is crossed, 60% of the RAID estimates had CEP measures of 76 meters or less compared with 60% of the staff estimates being measured as 115 meters or less.

Reading the plot up from the x-axis, one can see the fraction of estimates that were more accurate than a given value of CEP. For example, reading up from 50 meters until each of the regression lines is crossed, 9.6% of the staff estimates were 50 meters or less compared with 33.9% of the RAID estimates being 50 meters or less.

# 6 Conclusions

One can perform validation of a fusion system by comparing its output errors with errors produced by an accepted standard fusion device, such as a group of human experts. Challenges of this approach include assuring that both fusion systems receive exactly the same information, and devising a domain-specific metric for assessing the magnitude of errors. In spite of these challenges, the approach can be implemented in practice, as we demonstrated in this paper.

The key advantage of this approach is mitigation of variability in the content of the input available to the fusion system. For example, when intentional concealment and deception are important factors in the problem domain, the information content available to the fusion system can vary drastically from problem to problem. In such cases, the fusion system is best evaluated by comparing its performance to the performance of an accepted "gold standard."

The experimental comparative approach is also invaluable in guiding the process of system development. Deficiencies of a fusion system under development, and the root causes of the deficiencies, become clear.

It is worth mentioning here the practical outcomes of the fusion validation effort described in the paper. According to the publically available reports [18, 19], the United States Army has been in the process of adopting the RAID tool into its military intelligence systems and battle command systems.

In these applications, RAID is a tool for semi-automated generation of enemy estimates: its role is to anticipate the upcoming actions of the enemy, both before, and during the unfolding battle, in near real time. In domain-specific terms, RAID performs so-called enemy estimates in mission analysis and preparation, or running estimates during the execution of a mission. Such a capability involves elements of Level-2 and Level-3 information fusion and data mining.

As part of adapting RAID to the US Army requirements, the tool was tested in live-force (i.e., realistic wargames with real soldiers, combat vehicles, etc.) experiments conducted by the US Army in Fort Monmouth, NJ. There, a company commander and his subordinates used RAID while riding in a command and control vehicle, accessing RAID via the US Army battle command system called FBCB2 [20]. It was also used by the battalion staff at a workstation in the battalion tactical operations center.

Before and during a typical multi-hour mission, RAID continuously read information from FBCB2, such as enemy observations reports and reports of friendly positions. It also received input and assumptions from the battalion staff or the company commander. Then, combining such data, RAID generated running estimates of the enemy situation, e.g. likely ambushes, concealed enemy positions, routes of infiltration or retreat.

Another application considered for RAID is the US Army military intelligence fusion system called DCGS-A [21]. As of 2007, the reported plans were to integrate RAID technologies into an experimental version of DCGS-A, and to assess experimentally its value to users under variety of real-world operational conditions.